\pdfoutput=1

\documentclass[11pt]{article}

\usepackage{EMNLP2023}

\usepackage{times}
\usepackage{latexsym}
\usepackage{amsmath} 
\usepackage{tabularx}
\usepackage{multirow}
\usepackage{booktabs}

\usepackage[T1]{fontenc}

\usepackage[utf8]{inputenc}
\usepackage{geometry} 
\usepackage{caption} 
\usepackage{array} 
\usepackage{rotating} 

\usepackage{microtype}
\usepackage{xcolor}
\usepackage{inconsolata}
\usepackage{booktabs} 
\usepackage{amsfonts}
\usepackage{amssymb}
\allowdisplaybreaks
\usepackage{tabularray}

\title{Continual Learning with Dirichlet Generative-based Rehearsal}

\author{Min Zeng$^1$, \large{\textbf{Wei Xue$^1$, Qifeng Liu$^{1,2}$, Yike Guo$^1$}}\\
\\
    $^1$Hong Kong University of Science and Technology\\
    $^2$Hong Kong Institute of Science \& Innovation, Chinese Academy of Sciences
	    }

\begin{document}
\maketitle
\begin{abstract}
Recent advancements in data-driven task-oriented dialogue systems (ToDs) struggle with incremental learning due to computational constraints and time-consuming issues. Continual Learning (CL) attempts to solve this by avoiding intensive pre-training, but it faces the problem of catastrophic forgetting (CF). While generative-based rehearsal CL methods have made significant strides, generating pseudo samples that accurately reflect the underlying task-specific distribution is still a challenge.
In this paper, we present Dirichlet Continual Learning (DCL), a novel generative-based rehearsal strategy for CL. Unlike the traditionally used Gaussian latent variable in the Conditional Variational Autoencoder (CVAE), DCL leverages the flexibility and versatility of the Dirichlet distribution to model the latent prior variable. This enables it to efficiently capture sentence-level features of previous tasks and effectively guide the generation of pseudo samples.
In addition, we introduce Jensen-Shannon Knowledge Distillation (JSKD), a robust logit-based knowledge distillation method that enhances knowledge transfer during pseudo sample generation. Our experiments confirm the efficacy of our approach in both intent detection and slot-filling tasks, outperforming state-of-the-art methods.
\end{abstract}

\section{Introduction}
\begin{figure}[!t]
    \centering
    \includegraphics[width=70mm]{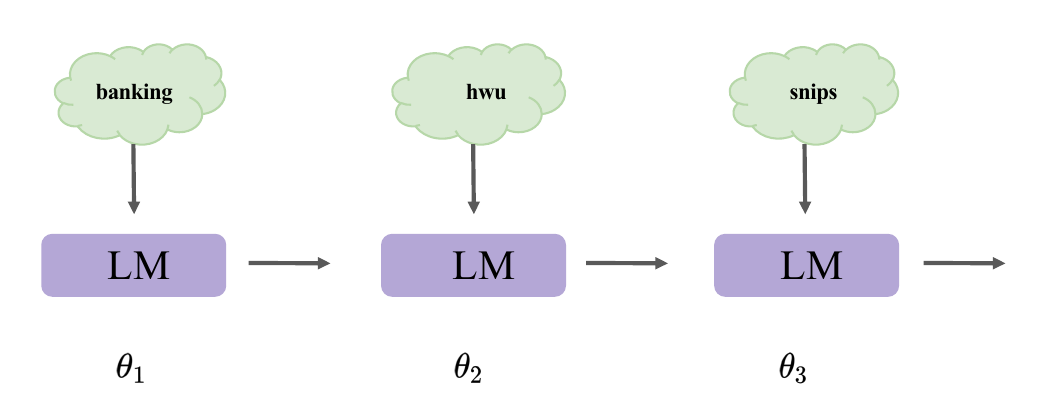}
    \caption{In this example of Continual Learning, the LM first trains on the \textit{banking} dataset, resulting in parameter $\theta_1$. The LM then trains on \textit{hwu}, followed by \textit{snips}, and so on. The parameters are updated sequentially.}
    \label{fig:cl}
    \vspace{-.5em}
\end{figure}

Large Language Models (LLMs) excel in many natural language processing (NLP) tasks, but they require significant resources, time, and data to train from scratch. Additionally, retraining them for each new task is often not feasible. To address these issues, continual learning (CL) is introduced. CL enables LLMs to learn new information from sequential tasks or datasets without losing their original performance. The process of CL is depicted in Figure \ref{fig:cl}.
Although CL performs better than direct fine-tuning, experiments consistently reveal the unavoidable problem of catastrophic forgetting (CF) \citep{mccloskey1989catastrophic}.
CF occurs when the performance on prior tasks deteriorates upon learning a new one, primarily due to shifts in data distribution between current and previous tasks.

Traditionally, the CL methods can be mitigated through three strategies: \emph{regularization}, \emph{architectural}, and \emph{rehearsal}. 
{\em Regularization} \citep{kirkpatrick2017overcoming, zenke2017continual, aljundi2018memory} minimally updates important parameters from previous tasks to retain performance, but the accumulating regularizers can over-constrain network parameters, hindering new task learning.
{\em Architectural} \citep{madotto2021continual, zhang2022continual} modifies the network structure for better task-specific feature extraction. Nevertheless, their individual task-focused approach may neglect knowledge transfer between old and new tasks.
{\em Rehearsal} \citep{lopez2017gradient,sunlamol, mi2020continual} utilizes episodic memory for task recall, with ``store-based rehearsal'' using stored real samples and ``generative-based rehearsal'' generating pseudo samples. The latter, being more memory-efficient, has received greater attention.

Generative-based rehearsal, including Prompt Conditioned VAE (CVAE) \citep{zhao2017learning} for Lifelong Learning (PCLL) \citep{zhao2022prompt}, has achieved state-of-the-art (SOTA) performance. 
Effective rehearsal hinges on the generative model's ability to closely mimic real samples from previous tasks. 
PCLL uses a symmetric Gaussian distribution to model discrete latent variables, often inaccurately capturing task-specific distributions. Additionally, VAE's Kullback–Leibler (KL) vanishing problems \citep{fu2019cyclical} lead to generating generic, similar samples.

In this paper, we target solving the CF problem of CL caused by distribution shifts in LLMs, aiming to preserve the domain-specific knowledge from streaming pre-training corpus distributions \citep{chen2023lifelong}. Our method builds on PCLL, L2KD \citep{chuang2020lifelong}, and Prompt Tuning \citep{zhu2022continual}, demonstrating that generative-based rehearsal mitigates CF without extra computational costs, while knowledge distillation and prompting enhance performance.
We propose Dirichlet Continual Learning (DCL), a new generative-based rehearsal method that combines task distribution modeling and knowledge distillation. Inspired by Latent Dirichlet Allocation (LDA) in topic modeling \citep{blei2003latent}, we treat NLP tasks as topics and employ a Dirichlet distribution-based CVAE for generating pseudo samples. Additionally, we introduce a logit-based knowledge distillation \citep{hinton2015distilling} method called Jensen–Shannon Knowledge Distillation (JSKD) under the CL knowledge distribution framework.
We evaluate our method in ToDs, and the results consistently show its superiority over the baselines. 
To summarize, our main contributions are:
\begin{itemize}
	\item We propose a Dirichlet distribution-based method to model the latent variable in CVAE, which achieves better rehearsal for continual learning of ToDs and mitigates catastrophic forgetting without access to the old data.
	\item We develop Jensen-Shannon Knowledge Distillation (JSKD), a new logit-based knowledge distillation strategy for knowledge transfer between teacher and student models.
	\item Experimental results in ToDs demonstrate that DCL improves baselines by a large margin.  
\end{itemize}

\section{Related Works}
\subsection{Continual Learning}
Continual learning involves three categories: \emph{regularization}, \emph{architectural}, and \emph{rehearsal}. 

\emph{Regularization} method, unlike L2 normalization that assigns the same weight to all model parameters, reinforces earlier knowledge by constraining crucial parameters and inserting a regularization term. Elastic Weight Consolidation (EWC) \citep{kirkpatrick2017overcoming} identifies and avoids updating important parameters, preserving previous task performance. Such strategy is seen in works like ARPER \citep{mi2020continual}, Progress \& Compress \citep{schwarz2018progress}, MAS \citep{aljundi2018memory}, and LwM \citep{Dhar_2019_CVPR}.

\emph{Architectural} approaches modify the network structure to reduce CF by adding task-specific parameters to the base models to effectively model task-specific features. Typical works include Progressive Neural Network \citep{rusu2016progressive}, Pathnet \citep{fernando2017pathnet}, AdapterCL \citep{madotto2021continual}, Piggyback GAN \citep{zhai2020piggyback}, and Semi-Supervised Lifelong Learning (SSLL) \citep{zhao2022semi}. For example, AdapterCL parameterizes each task using residual adapters.

\emph{Rehearsal} methods, which maintain performance by using samples from previous tasks, are divided into store-based and generative-based types. Store-based methods like ICaRL \citep{Rebuffi_2017_CVPR} and Gradient Episodic Memory (GEM) \citep{lopez2017gradient} use stored samples. ICaRL applied a herding-based step to choose representative samples, while GEM uses memory to avoid forgetting and encourage positive backward transfer. Generative-based methods like ReMix \citep{Mi_2020_CVPR_Workshops}, the method by \citet{shin2017continual}, and Prompt Tuning \citep{zhu2022continual} create pseudo samples. ReMix and \citet{shin2017continual} create samples with Mixup and GAN respectively, while Prompt Tuning uses task-specific prompts. PCLL \citep{zhu2022continual}, similarly, uses a CVAE to create samples of past tasks.

\subsection{Task-Oriented Dialogue Modelling}
\label{sec:2.2}

Dialogue systems are split into Task-Oriented Dialogue Systems (ToDs) \citep{williams2007partially, wen2016network} and Chit-Chat Dialogue Systems (CcDs) \citep{shang2015neural,serban2016building}. ToDs perform specific tasks (like hotel booking), while CcDs offer non-targeted dialogue for psychological support. ToDs modules include Speech Recognition (SR), Natural Language Understanding (NLU), Dialogue Management (DM), and Natural Language Generation (NLG). NLU, a key module, interprets utterance knowledge and includes domain identification, intent detection, and slot filling. For a hotel booking task, domain identification identifies the topic (i.e., the hotel), intent detection identifies the booking request by recognizing the user's intention in utterance and slot filling returns hotel information like name and address. The DM module updates the global dialogue state using Dialogue State Tracking (DST) and applies Dialogue Policy to determine system actions. NLG \citep{press2017language} then generates dialogue responses.

\subsection{Latent Dirichlet Allocation}
Latent Dirichlet Allocation (LDA) \citep{blei2003latent}, a popular topic model, uses the Dirichlet distribution to model topic and word distributions in documents. It serves as the conjugate prior to the multinomial distribution. LDA-based document models for ad-hoc retrieval were proposed in \citep{wei2006lda}. An online variational bayes (VB) algorithm for LDA was developed by \citet{hoffman2010online}, and \citet{foulds2013stochastic} propose a stochastic algorithm for collapsed VB inference in LDA. The Embedded Topic Model (ETM) \citep{dieng2020topic} combines LDA and word embeddings to identify interpretable topics with large vocabularies, including rare and stop words. In addition, \citet{li2020dirichlet} introduce a Dirichlet graph VAE for graph generation and clustering.

\begin{figure*}[!t]
    \centering    \includegraphics[width=130mm]{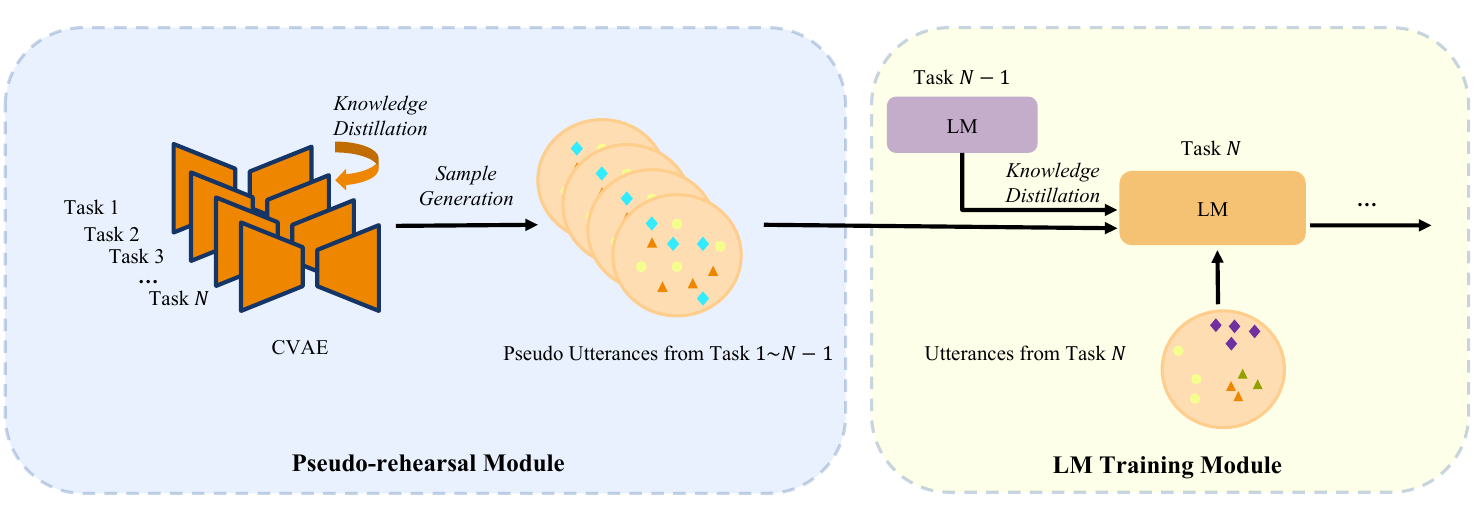}
    \caption{Overview of the proposed DCL model. DCL consists of two modules: the pseudo-rehearsal module and the LM training module. The pipeline of DCL can be summarized as follows: (1) In task $N$ training, the pseudo-rehearsal module uses a CVAE to produce pseudo samples from tasks $1$ to $N-1$. (2) These pseudo samples are then mixed with the dataset of task $N$ and used in the current task training in the LM training module.}
    \label{fig:overview1}
    \vspace{-.5em}
\end{figure*}

\section{Methodology}


\subsection{Task Definition}
\label{sec:def}
In this paper, we focus on the NLU of ToDs which learns the feature and knowledge from the utterance and generally includes domain identification, intent detection, and slot filling. Here following the common practice \citep{zhao2022prompt, madotto2021continual, mi2020continual, zhu2022continual} and to facilitate comparison, we mainly consider intent detection and slot filling tasks. For learning the tasks of ToDs in the CL manner, we assume that there is a sequence of tasks $T=\{T_1, \cdots, T_N\}$, and an LM model which is expected to solve the tasks by gradually training on the samples from the sequence of tasks. Examples of intent detection and slot filling have been explained in Section \ref{sec:2.2}.

\subsection{Overview}
\label{sec:archi}



In CL, the quality of exemplars is critical for preserving previous task performance, as emphasized in \citep{mi2020continual}. Thus, generating representative and diverse pseudo utterances is important. Among commonly used generative models, VAE and GAN are prominent. VAE, in particular, is known for generating meaningful, diverse dialogues \citep{serban2017hierarchical}. Therefore, we propose a CVAE-based DCL that utilizes latent variables.

DCL, shown in Figure \ref{fig:overview1}, has two main modules: pseudo-rehearsal and Language Model (LM) training. The pseudo-rehearsal module employs CVAE to generate pseudo samples from previous tasks. Then, the LM is updated using both these samples and those from the current task, enabling CL. The total training loss of DCL combines the CVAE loss for pseudo-rehearsal and the LM loss for updating LM parameters.



The structure of the CVAE and LM models is outlined as follows. In CVAE, both encoder and decoder employ GPT-2 \citep{radford2019language} with distinct parameters to encode information and generate pseudo samples for tasks. In a continual learning context, the LM is expected to generate samples for task sequences, making the decoder of the CVAE also function as the LM as depicted in the LM training module of Figure \ref{fig:overview1}. Consequently, the same model updating and cross-task knowledge distillation processes are applied to both the CVAE and LM models.

 

The key of the DCL is to generate pseudo samples. In the proposed DCL model, we replace the Gaussian latent with the Dirichlet latent and employ reject sampling \citep{jankowiak2018pathwise} to reparametrize the Dirichlet latent variable.

\subsection{Dirichlet-guided Pseudo-rehearsal Module}
\label{sec:3.3}
This module aims to generate the pseudo samples for rehearsal based on the task ID. Exploiting the 
modeling flexibility of the Dirichlet distribution, we treat different NLP tasks as topics and propose a Dirichlet-guided pseudo-rehearsal module. 

For the task $T_n$, we expect to generate $y_i$ given the input utterance $x_i$. To achieve task-dependent generation, a specific prompt ${Pr}_n$ for $T_n$ is first defined, and then concatenated to the input utterance, yielding the augmented input $\Tilde{x}_{i,n}={Pr}_n \oplus x_i$. Further, the CVAE can be utilized to generate the pseudo samples for $T_n$ based on $\Tilde{x}_{i,n}$ \citep{zhao2017learning}. The key idea of CVAE is to reconstruct the input $x$ through the latent variable $z$, which is normally modeled through the Gaussian distribution. 

The CVAE is trained to maximize the log-likelihood $\log {p_\theta(x)}$, where $\theta$ is the model parameter. However, $\log {p_\theta(x)}$ is intractable \citep{kingma2013auto}, the lower bound ELBO $\mathcal{L}(\theta,\phi;x,c)$ is used for tractable optimization:
\begin{align} 
\small
\label{elbo}
\log{p_\theta(x)}
& \ge \mathcal{L}(\theta,\phi;x,c)\nonumber\\
&=-\lambda KL(q_\phi(z|x, c)||p_\theta(z|c)) \nonumber\\
&\quad+\mathbb{E}_{q_\phi(z|c, x)}[\log{p_\theta}(x|z, c)]\nonumber\\
&\le \log{p(x|c)},
\end{align}
where $p_\theta(z|c)$ is the prior distribution of $z$, $q_\phi(z|x, c)$ approximates the intractable true posterior distribution, $c$ defines the task ID, and $\lambda$ is the dynamic KL weight to mitigate the KL-vanishing, as proposed by \citet{bowman2016generating}. 

However, as illustrated in \citep{shen2018improving, zeng2019dirichlet}, although the weighting scheme can be used, KL-vanishing can not be essentially tackled. The main reason is that a symmetric Gaussian from continuous space is not flexible and sufficient enough to express the latent $z$ originating from discrete space. Here, we introduce the Dirichlet distribution, which uses a more flexible structure to approximate the prior distribution of $z$. The versatile forms of the Dirichlet distribution, which can be concave, convex, symmetrical, or asymmetrical, make it an appealing choice for our model.

The CVAE loss denoted as $\mathcal{L}_{\rm CVAE}$ is the negative of ELBO. Following \eqref{elbo}, we have
\begin{equation}
\mathcal{L}_{\rm CVAE} = \mathcal{L}'_{\rm KL}+\mathcal{L}_{\rm Rec},
\end{equation}
and $\mathcal{L}'_{\rm KL}$ can be expressed as follows after derivation \citep{zeng2019dirichlet}:
\begin{align}
\small
\label{derivation_kl}
&KL(q_\phi(z|x,c)||p(z|c))= \nonumber\\
&\log\Gamma(\sum_{k=1}^{K}\alpha_k)-\sum_{k=1}^{K}\log\Gamma(\alpha_k)\nonumber\\
&-\log\Gamma(\sum_{k=1}^{K}\beta_k)
+\sum_{k=1}^{K}\log\Gamma(\beta_k)\nonumber\\
&+\sum_{k=1}^{K}(\alpha_k-\beta_k)(\psi(\alpha_k)-\psi(\sum_{k=1}^{K}\alpha_k)),
\end{align}
where $\alpha$ and $\beta$ represent the parameters of the Dirichlet distributions $q_\phi(z|x,c)$ and $p_\theta(z|c)$, respectively, $K$ denotes the dimension of $z$, and $\psi$ is the Digamma function.

\subsection{LM Training Module}
\label{sec:lm-module}
LM module uses GPT-2 for training. Taking task $T_n$ for example, the pseudo-rehearsal module first generates pseudo samples of previous tasks $T_1, \cdots, T_{n-1}$, then we combine the generated pseudo samples with the samples for the task $T_n$ to update the model. Hence, training dataset for task $T_n$ becomes $\mathcal{D}_{cups}=\mathcal{D}_{curr}\cup\mathcal{D}_{pseu}$. The training loss is defined as: 
\begin{align}
\small
&\mathcal{L}_{\rm LM}(\theta)\nonumber\\
&=-\sum_{(x_i,y_i)\in \mathcal{D}_{cups}}^{}\log p_{\theta}{(x_i,y_i)} 
+ \log p_{\theta}{(y_i|x_i)}.
\end{align}

\subsection{Jensen-Shannon Knowledge Distillation}
\label{sec:js}

\begin{figure}[!t]
    \centering
    \includegraphics[width=70mm]{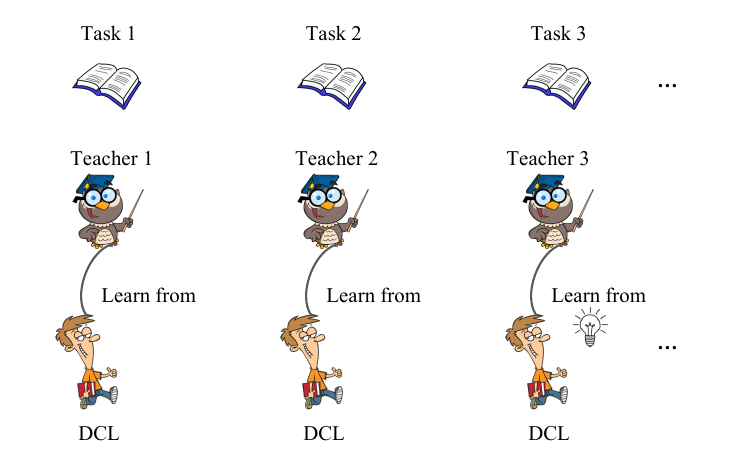}
    \caption{Knowledge Distillation of DCL.}
    \label{fig:kd}
    \vspace{-.8em}
\end{figure}

We further propose Jensen-Shannon Knowledge Distillation (JSKD) to help the model remember previous tasks. Many CL studies \citep{chuang2020lifelong, mi2020continual, zhao2022prompt, chen2023lifelong} use knowledge distillation to lessen CF. Like L2KD \citep{chuang2020lifelong}, as illustrated in Figure \ref{fig:kd}, DCL starts training on a new task with a teacher model, then transfers the knowledge to a student model. In our case, the teacher model is trained on the old task and the student model on the new task. This helps the CL model adapt to the new task while retaining the knowledge from previous tasks simultaneously. We now explain JSKD and compare it with the traditional KL-based knowledge distillation method.

\noindent\textbf{Knowledge Distillation}
Given a training sample $(x,y)$, we want to minimize the cross-entropy between the output distribution of the teacher and student models. The training objective is:
\begin{equation}
\small
\mathcal{L}_{\rm KD} = \alpha \cdot \mathcal{L}_{\rm KL}(S, T) \cdot\tau^2  + (1 - \alpha) \cdot \mathcal{L}_{\rm CE}(S, Y),
\end{equation} 
where $T$ and $S$ are teacher and student predictions, respectively. $\tau$ is the temperature to soften the teacher's predictions, while $\mathcal{L}_{\rm CE}(S, Y)$ quantifies the cross-entropy loss between student predictions and the ground truth labels $Y$. $\mathcal{L}_{\rm KL}$ implicitly prevents the student's model parameters from straying too far from the teacher's model parameters. Also, the first term denotes a soft target while the second term is the hard target. $\alpha\in[0, 1]$ balances the soft and hard target evaluations.

\noindent\textbf{JS Divergence vs KL Divergence} For distributions $p$ and $q$, JS divergence \citep{lin1991divergence} is defined by :
\begin{align}
\small
\mathcal{L}_{\rm JS}(p \parallel q) = &\frac{1}{2} {\mathcal{L}_{\rm KL}\left(p \parallel \frac{1}{2}(p + q)\right)} \nonumber\\
&+ \frac{1}{2} {\mathcal{L}_{\rm KL}\left(q \parallel \frac{1}{2}(p + q)\right)}.
\end{align}
JS divergence symmetrically measures the similarity between two probability distributions, in contrast to the asymmetric KL divergence, with values ranging from 0 (identical distributions) to 1(no shared support). JS divergence offers advantages over KL divergence: a) Its symmetry ensures consistent values regardless of comparison order, making it ideal for measuring distribution similarities. b) JS provides bounded value in $[0,1]$, while KL divergence spans $[0,+\infty)$.  
The above properties make JS divergence more suitable for knowledge distillation than KL since the KL divergence will be infinite when one sample appears only in one task distribution.

\noindent\textbf{Knowledge Distillation via JS Divergence }
Motivated by the above discussions, we propose a JS divergence-based knowledge distillation (JSKD) to more accurately measure the distance between teacher and student models, enhancing model robustness. The JSKD loss is defined as:
\begin{equation}
\small
\label{kd}
\mathcal{L}_{\rm KD} = \alpha \cdot \mathcal{L}_{\rm JS}(S, T) \cdot \tau^2 + (1 - \alpha) \cdot \mathcal{L}_{\rm CE}(S, Y). 
\end{equation} 
Specifically, we use the preceding task for the teacher model and the current task for the student model. As mentioned, DCL optimizes $\mathcal{L}_{\rm CVAE}+ \mathcal{L}_{\rm LM}$. For CVAE, the training loss
$\mathcal{L}_{\rm CVAE}=\mathcal{L}_{\rm Rec} + \mathcal{L}'_{\rm KL}$. 
Incorporating knowledge distillation, the $\mathcal{L}_{\rm Rec}$ and $\mathcal{L}_{\rm LM}$ for task $T_n$ is defined as:
\begin{align}
&\mathcal{L_{\rm Rec}} = \alpha\mathcal{L}_{\rm JS}(l_c, l_c^*)\tau^2  + (1 - \alpha) \mathcal{L}_{\rm CE}(l_c, Y),\nonumber\\
&\mathcal{L}_{\rm LM} = \alpha\mathcal{L}_{\rm JS}(l_l, l_l^*) \tau^2 + (1 - \alpha) \mathcal{L}_{\rm CE}(l_l, Y),
\end{align}
where $l_c$ and $l_l$ are the logits output of CVAE and LM of task $T_n$, respectively. $l_c^*$ and $l_l^*$ represent the logits output of task $T_{n-1}$, and $Y$ signifies the ground truth. We emphasize that $\mathcal{L}'_{\rm KL}$ is the KL loss in \eqref{derivation_kl} to evaluate the distance between the assumed Dirichlet data distribution and the real distribution. It is different from the $\mathcal{L}_{\rm KL}$ for evaluating the distance between the student and teacher models in cross-task knowledge distillation.

\section{Experiments}
\subsection{Datasets}
We evaluate our proposed model by using distinct datasets for two separate tasks: intent detection and slot filling.  
For intent detection, we employ HWU \citep{liu2021benchmarking}, BANKING \citep{casanueva2020efficient}, CLINC \citep{larson2019evaluation}, SNIPS \citep{coucke2018snips}, AITS \citep{hemphill1990atis}, and TOP \citep{gupta2018semantic} datasets. 
Consistent with previous works \citep{zhao2022prompt}, we divide the TOP dataset into three separate subsets: TOP-S1, TOP-S2, and TOP-S3. Each is treated as an individual task to expand the number of tasks for CL evaluation.
For slot filling, SNIPS, AITS, DSTC \citep{rastogi2020towards}, MIT-MOVIE \footnote{\url{https://groups.csail.mit.edu/sls/downloads/}\label{mit-courpus}}, and MIT-RESTAURANT \footref{mit-courpus} datasets are used. 

For a fair comparison, these tasks are learned in six different orders, and the average performances across these orders are reported.

\subsection{Baselines}
To demonstrate the effectiveness of our approach, we compare it with eleven robust baselines. We note that we \textbf{Fine-tune} the pre-trained language models GPT2 \citep{radford2019language} on the stream of tasks without any strategy to prevent CF. We use multi-task (\textbf{Multi}) learning as the upper bound. 

 \noindent\textit{\textbf{Regularization:}} \textbf{EWC} \citep{kirkpatrick2017overcoming} is a regularization method that mitigates catastrophic forgetting by constraining crucial parameters while enabling less significant ones adapted to new-task data. \textbf{MAS} \citep{aljundi2018memory} quantifies parameter importance in the network based on task memory contributions, aiding in mitigating CF. 

\noindent\textit{\textbf{Rehearsal:}} \textbf{LAMOL} is a rehearsal method that utilizes the language model as both learner and generator, facilitating the creation of pseudo samples for current training. Its variations, \textbf{LAMOL-g} and \textbf{LAMOL-t}, diverge in terms of the incorporation of global or task-specific tokens. \textbf{L2KD} \citep{chuang2020lifelong} is built upon LAMOL which is proposed to introduce knowledge distillation into LAMOL. \textbf{ER} \citep{rolnick2019experience} uses on-policy learning for quick adaptation to new tasks and off-policy learning with behavioral cloning to enhance the performance of past tasks. \textbf{PCLL} \citep{zhao2022prompt} is a CVAE-based generative replay method that reaches the SOTA performance in this setting. 

\noindent\textit{\textbf{Architectural:}} \textbf{HAT} \citep{serra2018overcoming} proposes a task-based hard attention mechanism that preserves information from previous tasks without affecting the learning of the current task.
\textbf{CTR} \citep{ke2021achieving} inserts a continual learning plug-in module in two locations in BERT \citep{devlin2019bert} to achieve both CF mitigation and knowledge transfer. \textbf{AdapterCL} \citep{madotto2021continual} leverages task-specific residual adapters in a frozen GPT-2 backbone, thereby reducing parameter number and promoting efficient continual learning.

\subsection{Experimental Settings}
All experiments are conducted on NVIDIA A100 GPU. Experimental settings are summarized as: 
(1) In intent detection, the batch size is 32 with a learning rate of 5e-5 and a pseudo sample rate of 0.2. The dimension of $z$ is 128, and we use the Adam optimizer. We set the maximum context length as 256 and train it for 5 epochs. 
(2) The differences between slot filling with the above intent detection settings include a) the dimension of $z$ is 512, b) the maximum context length is 50, and c) we train it for 10 epochs.

\subsection{Evaluation Metrics}

\noindent\textbf{Average Joint Goal Accuracy (JGA):} Average JGA denotes the average accuracy on all tasks after the final task has been learned, which is defined as:
    ${\rm Avg.JGA} = \frac{1}{T} \sum\limits_{i=1}^{T}{R_{T, i}}$,
where $R_{i, j}$ denotes the evaluation metric on task $t_j$ after training on task $t_i$. Since intent detection and slot filling can be viewed as classification and sequence labeling tasks, we usually adopt accuracy (ACC) and F1 score (F1) for intent detection and slot filling, respectively.  

\noindent\textbf{Learning Curve Area (LCA):} We also use LCA \citep{chaudhryefficient}, computed as the area under a learning curve, to indicate a model's CL performance over a sequence of tasks. LCA is defined as:
${\rm LCA} = \int_{0}^{T} P(t) dt$,
where $P(t)$ is the average model performance at step $t$ across all already-learnt tasks, and $T$ is the total number of steps. Higher LCA values suggest efficient CL.

\section{Results and Analysis} 
\subsection{Overall Evaluation Results} 
Table~\ref{tab:overall} summarizes the performances of different methods, providing compelling evidence that our model outperforms all baselines. Superior results from our method suggest better pseudo sample generation and more effective knowledge transfer.
In particular, compared to the SOTA model PCLL, DCL shows a significant increase of $3.48\%$ in accuracy and $4.22\%$ in LCA for intent detection. Moreover, we observe an improvement of $2.89\%$ in F1 and $6.07\%$ in LCA for slot filling. These results indicate that using Dirichlet-guided pseudo-rehearsal and JSKD can mitigate CF.
Notably, the performance of our model is near the upper bound (multi-task learning), with only a small gap of $2.52\%$ in accuracy for intent detection, and a $3.43\%$ difference in F1 for slot filling.

\begin{figure}[!t]
    \centering
    \includegraphics[width=75mm]{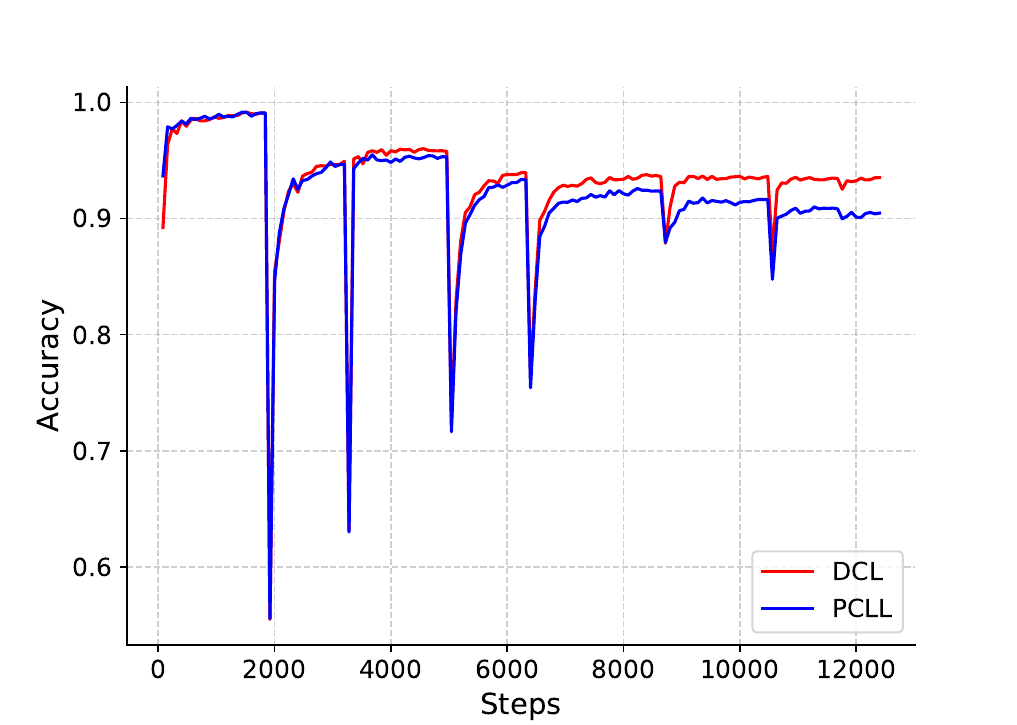}
    \caption{Accuracy of DCL and PCLL on intent detection task.}
    \label{fig:curve}
    \vspace{-.8em}
\end{figure}

To further understand these trends, we plot the learning curve of the average scores for DCL and PCLL in intent detection tasks in Figure \ref{fig:curve}. It is evident that our model alleviates the CF problem more effectively than the current SOTA in the continual learning process. The observed drop in accuracy is due to task switching.
We further interpret the result from two perspectives:
\begin{itemize}
    \item DCL significantly outperforms PCLL, suggesting that the Dirichlet latent variable approximates the data distribution more accurately. This leads to higher quality pseudo samples and overall better performance.
    \item Even though DCL generates fewer pseudo samples compared to the real samples used in multi-task learning (our upper bound), it still delivers strong performance. This implies that the pseudo samples generated by DCL are diverse and representative enough to capture the information present in the real samples.
\end{itemize}

\begin{table}[!t]
\centering
\resizebox{1\linewidth}{!}{
\begin{tabular}{lccccc}
\toprule
\multicolumn{1}{c}{\multirow{2}{*}{\textbf{Models}}} & \multicolumn{2}{c}{\textbf{Intent Detection (\%)}}  & \multicolumn{2}{c}{\textbf{Slot Filling (\%)}}                 \\
& \multicolumn{1}{c}{\textbf{ACC $\uparrow$}} & \multicolumn{1}{c}{\textbf{LCA $\uparrow$}}  & \multicolumn{1}{c}{\textbf{F1 $\uparrow$}} & \multicolumn{1}{c}{\textbf{LCA $\uparrow$}}
\\ \hline
Finetune  &14.09 & 28.76 & 15.38&19.55\\
EWC & 14.16 & 28.34 & 15.67&19.51\\
MAS & 14.15 & 28.61 & 15.59&19.37 \\
L2KD & 35.22 & 61.78 & 44.16&39.94\\
LAMOL-g & 50.30 & 60.67 & 45.12 & 38.03 \\
LAMOL-t & 51.81 & 67.97 & 44.83 & 37.58\\
ER & 78.19 & 78.19  & 44.95 & 39.32\\
HAT &73.92 & 73.03  & 61.99 & 67.33\\
CTR & 67.44 & 71.11 & 63.84 & 67.28\\
AdapterCL & 81.15 & 75.60 & 75.60 &48.47 \\
PCLL & 90.25 & 88.82 & 74.48 & 68.41\\
Multi (Upper Bound) & 96.25 & N/A & 80.80 & N/A\\
\hline
\textbf{DCL}  & \textbf{93.73} & \textbf{93.04} & \textbf{77.37} & \textbf{74.49}\\
\bottomrule
\end{tabular}
}
\caption{Evaluation results for DCL and baselines. 
Baseline scores are reported in \citep{zhao2022prompt} and multi-task performance is the upper bound. The best results are highlighted in bold.}
\label{tab:overall}
\vspace{-.8em}
\end{table}

\subsection{Ablation Study}


\noindent\textbf{Dirichlet or Gaussian-guided Rehearsal Module.} To understand the impact of using a Dirichlet-guided rehearsal module, we compared the performance of DCL and PCLL in intent detection and slot filling tasks, both using KL knowledge distillation. The distinction lies in the choice of either Dirichlet or Gaussian latent variables. Table \ref{tab:dir-gau} reveals that DCL with a Dirichlet-guided rehearsal module surpasses PCLL which uses a Gaussian-guided module, suggesting that the Dirichlet distribution is more effective at approximating the true data distribution.

\begin{table}[!t]
  \resizebox{1\linewidth}{!}{
  \begin{tabular}{lccccc}
\toprule
\multicolumn{1}{c}{\multirow{2}{*}{\textbf{Models}}} & \multicolumn{2}{c}{\textbf{Intent Detection (\%)}}  & \multicolumn{2}{c}{\textbf{Slot Filling (\%)}}                 \\
& \multicolumn{1}{c}{\textbf{ACC $\uparrow$}} & \multicolumn{1}{c}{\textbf{LCA $\uparrow$}}  & \multicolumn{1}{c}{\textbf{F1 $\uparrow$}} & \multicolumn{1}{c}{\textbf{LCA $\uparrow$}}
\\ \hline
   PCLL (with KL) & 90.25 & 88.82 & 74.48 & 68.41\\
   DCL (with KL)  & 92.83 & 91.32 & 76.42 & 73.76 \\
   \bottomrule
  \end{tabular}}
    \caption{Results of DCL and PCLL with KL knowledge distillation.}
    \label{tab:dir-gau}
\end{table}


\noindent\textbf{JS or KL Knowledge Distillation.} Next, we evaluate the impact of JS Knowledge Distillation. Table~\ref{tab:ablation-kl} shows the performance differences between DCL implementations that use either KL or JS knowledge distillation in the slot-filling task for various task learning orders. The results indicate that the model using JS Knowledge Distillation outperforms the one using KL knowledge distillation. This suggests that JS divergence is more effective for knowledge transfer.

\begin{table}[t]
\small
\centering
\renewcommand{\arraystretch}{1} 
\resizebox{0.8\linewidth}{!}
{
\begin{tabular}{lccccc}
\toprule
\multicolumn{1}{c}{\multirow{2}{*}{\textbf{Orders}}} & \multicolumn{2}{c}{\textbf{DCL (with KL)}}  & \multicolumn{2}{c}{\textbf{DCL (with JS)}}                 \\
& \multicolumn{1}{c}{\textbf{F1} $\uparrow$} & \multicolumn{1}{c}{\textbf{LCA} $\uparrow$}  & \multicolumn{1}{c}{\textbf{F1} $\uparrow$} & \multicolumn{1}{c}{\textbf{LCA} $\uparrow$}
\\ \hline
order 0 & 80.26 & 73.82 & 81.39 & 74.66\\
order 1 & 80.31 & 74.82 & 80.94 & 75.23\\
order 2 & 74.04 & 70.01 & 74.77 & 70.73 \\
order 3 & 77.00 & 76.28 & 78.03 & 76.69\\
order 4 & 72.98 & 70.10 & 74.53 & 70.99 \\
order 5 & 73.92 & 77.50  & 74.56 & 78.61 \\
\hline
Mean & 76.42 & 73.76 & 77.37 & 74.49\\
\bottomrule
\end{tabular}
}
\caption{Slot filling results of F1 (\%) and LCA (\%) on six orders with KL and JS knowledge distillation.}
\label{tab:ablation-kl}
\vspace{-.8em}
\end{table}


\noindent\textbf{Number of Pseudo Samples.}
We conducted further analysis to understand how the number of pseudo samples impacts the performance of the proposed approach, by testing various pseudo sample ratios in DCL. Table \ref{tab:ablation-psn} presents the experimental results for pseudo sample ratios of 0.1, 0.2, 0.4, and 0.5. Even when fewer pseudo samples are added to the training, DCL with a ratio of 0.1 still outperforms PCLL with a ratio of 0.2. Moreover, we found that increasing the number of pseudo samples further improves performance. This is expected because more data samples contain more information, enhancing the model's capabilities.

\begin{table}[h]
\small
\centering
{
\begin{tabular}{lccc}
\toprule
\textbf{Ratio} &\textbf{ACC} $\uparrow$ & \textbf{LCA} $\uparrow$
\\ \hline
0.1 & 91.66 & 91.83 \\
0.2 & 93.73 &93.04\\
0.4 & 93.97 & 92.76 \\
0.5 & 94.23 & 92.82  \\
\bottomrule
\end{tabular}
}
\caption{Intent detection result of DCL with a different number of pseudo samples numbers added in.} 
\label{tab:ablation-psn}
\end{table}



\noindent\textbf{Evaluating Pseudo Samples Quality.}
We establish that the quality of DCL's pseudo samples surpasses that of the baselines, as evidenced by the higher accuracy and LCA. Notably, our method outperforms PCLL, which differs only in its use of Gaussian latent variable as opposed to our proposed Dirichlet variables. This suggests that Dirichlet latent variable are more effective than Gaussian ones in generating pseudo samples.
We utilize  \textbf{Dist-n} \cite{li2015diversity} to assess the quality of pseudo samples. Dist-n measures the proportion of distinct n-grams in the generated pseudo samples. A higher Dist-n value, indicating greater pseudo-sample diversity, is generally preferred as it shows the samples are more varied.
Given the limited number of pseudo samples included, the quality of our exemplars is crucial in preserving the performance of previous tasks. Our aim is to carefully select representative and diverse utterances, rather than settling for generic and closely similar ones.

Table~\ref{tab:ablation-dist} summarizes the Dist-n results. Notably, DCL achieves higher distinct scores compared to other methods, indicating that DCL-generated pseudo samples exhibit greater diversity. This suggests that pseudo samples created using DCL more closely resemble real samples.

\begin{table}[t]
\centering
\renewcommand{\arraystretch}{1} 
\resizebox{0.9\linewidth}{!}
{
\begin{tabular}{lccccc}
\toprule
\textbf{Models} & \textbf{Dist-1} & \textbf{Dist-2} & \textbf{Dist-3} & \textbf{Dist-4}
\\ \hline
LAMOL-g & 0.0602 & 0.2466 & 0.4489 & 0.6178\\
LAMOL-t & 0.1758 & 0.4733 & 0.6837 & 0.8090\\
PCLL & 0.2836 & 0.6566 & 0.8369 & 0.9221 \\
\textbf{DCL} & \textbf{0.3092} & \textbf{0.7019} & \textbf{0.8708} & \textbf{0.9389} \\
\hline
Real Sample & 0.4000 & 0.7972 & 0.9255 & 0.9717\\
\bottomrule
\end{tabular}
}
\caption{Distinct scores for generated pseudo samples.} 
\label{tab:ablation-dist}
\vspace{-.8em}
\end{table}



\noindent\textbf{Dimension of Latent Variable.}
We also examine the impact of the latent variable $z$'s dimension, as displayed in Table~\ref{tab:ablation-dim-z}. It shows that DCL using JSKD with a latent dimension of $8$ outperforms DCL using KL with a dimension of $128$. This suggests that the Dirichlet latent is superior to the Gaussian latent. Even with smaller dimensions and less information encoded, the model can generate high-quality pseudo samples, resulting in improved accuracy. 
However, it's worth noting that DCL using JSKD with a latent dimension of $8$ doesn't perform as well as DCL with a dimension of $128$. This can be attributed to the reduced information capacity of the smaller $z$ dimension, which may lead to a decrease in performance.

\begin{table}[t]
\small
\centering
 {
\begin{tabular}{lccc}
\toprule
\textbf{Models} & \textbf{ACC} $\uparrow$ & \textbf{LCA} $\uparrow$
\\ \hline
DCL (with KL, z = 128)  & 92.83 & 91.32 \\
DCL (with JS, z = 8)  & 93.51 & 93.11 \\
DCL (with JS, z = 128)  & 93.73 & 93.04 \\
   \hline
  \end{tabular}}
    \caption{Intent detection result of DCL (with JS) and DCL (with KL) for different latent variable dimensions.}
    \label{tab:ablation-dim-z}
    \vspace{-.3em}
\end{table}


\begin{table}[ht]
\small
\centering
\begin{tabularx}{\columnwidth}{
  >{\hsize=.4\hsize\raggedright\arraybackslash}X
  >{\hsize=1.8\hsize\raggedright\arraybackslash}X
  >{\hsize=.8\hsize\raggedright\arraybackslash}X
}
\toprule
\textbf{Models}  & \textbf{Input Utterance} & \textbf{Output y} \\
\midrule
Golden & {1. What's the fuel economy of my car.} & { 1. mpg} \\
   & { 2. What is the expiration date on my card?} & { 2. expiration date} \\
\midrule
PCLL & {1. Do they have a lot of miles on this road?} & { 1. mpg} \\
   & { 2. Do you know how much my new credit card is worth?} & { 2. expiration date} \\
\midrule
DCL & {1. What is the mpg of this car?} & { 1. mpg} \\
   &{2. Can you check my expiration month?} & { 2. expiration date} \\
\bottomrule
\end{tabularx}
\caption{Comparison of Generated Pseudo Samples by PCLL and DCL against the Ground Truth (Golden).}
\label{tab:casestudy}
\vspace{-.8mm}
\end{table}

\subsection{Case Study}
Table~\ref{tab:casestudy} presents a comparison between the pseudo samples generated by both DCL and PCLL and real samples, and examples generated on CLINC from the intent detection task are shown. A pseudo sample includes the input utterance (middle column) as well as the intent (right column). It is obvious that PCLL struggles to generate the intent of specific sentences correctly. For instance, PCLL wrongly generates the intent \textit{``mpg''} (miles per gallon) for the utterance \textit{``Do they have a lot of miles on this road''}, showing that PCLL fails to capture the actual meaning of the utterance. 
In addition, for the utterance \textit{``Do you know how much my new credit card is worth?''}, PCLL also wrongly detects the intent as the \textit{``expiration date''} which is actually not relevant to the input. 



\section{Conclusions}
In this paper, we propose DCL, a generative-based rehearsal method to alleviate catastrophic forgetting for continual learning in ToDs. 
A Dirichlet distribution-based CVAE is developed to exploit the flexibility of Dirichlet distribution to model the utterance-level characteristics, 
improving pseudo sample generation compared to the traditional Gaussian-based CVAE.
We also proposed a more robust JS divergence-based knowledge distillation method to facilitate knowledge transfer between tasks. Comprehensive experiments show the superiority of the proposed method.

\newpage
\section{Limitations}
The limitations of our work include:
\begin{itemize}
    \item Our model can be further mixed with the architectural method for better performance. For example, a Dirichlet latent variable can be introduced to grasp the global characteristics for a specific task. Consequently, task-specific residual adapters in the LM training module can be designed to capture each task's local features. 
    \item We infer from Table \ref{tab:ablation-dim-z} that many dimensions of $z$ are inactive for the final performance. The interpretable relation between the dimensions and the final performance is not investigated, which could potentially help to achieve controlled generation.
\end{itemize}





\bibliography{custom}
\bibliographystyle{acl_natbib}


\end{document}